\def\year{2021}\relax
\documentclass[letterpaper]{article} % DO NOT CHANGE THIS
\usepackage{aaai21}  % DO NOT CHANGE THIS
\usepackage{times}  % DO NOT CHANGE THIS
\usepackage{helvet} % DO NOT CHANGE THIS
\usepackage{courier}  % DO NOT CHANGE THIS
\usepackage[hyphens]{url}  % DO NOT CHANGE THIS
\usepackage{graphicx} % DO NOT CHANGE THIS
\usepackage{natbib}  % DO NOT CHANGE THIS AND DO NOT ADD ANY OPTIONS TO IT
\usepackage{caption} % DO NOT CHANGE THIS AND DO NOT ADD ANY OPTIONS TO IT
\usepackage{multirow}
\usepackage{amsthm,amsmath,amssymb}
\usepackage{mathrsfs}
\usepackage{color}
\usepackage{pifont}    
\usepackage{makecell}
\newcommand{\etal}{\textit{et al. }}
\newcommand{\eg}{\textit{e.g., }}
\newcommand{\etc}{\textit{etc}}
\newcommand{\ie}{\textit{i.e., }}

\title{SPIN: Structure-Preserving Inner Offset Network for Scene Text Recognition}
\author{Chengwei Zhang\textsuperscript{\rm 1}\footnotemark[1]
Yunlu Xu\textsuperscript{\rm 2}\thanks{Authors contribute equally. Zhang did this work when he was an intern in Hikvision Research Institute.}
Zhanzhan Cheng\textsuperscript{\rm 32}\thanks{This work is completed under the supervision of Zhanzhan Cheng (contact email: chengzhanzhan@hikvision.com).}
Shiliang Pu\textsuperscript{\rm 2}\footnotemark[3] 
Yi Niu\textsuperscript{\rm 2} Fei Wu\textsuperscript{\rm 3} Futai Zou\textsuperscript{\rm 1}\thanks{Corresponding author.}\\}
\affiliations{
\textsuperscript{\rm 1}Shanghai Jiaotong University, China; ~~~\textsuperscript{\rm 2}Hikvision Research Institute, China; ~~~\textsuperscript{\rm 3}Zhejiang University, China\\
\{cwzhang,~zoufutai\}@sjtu.edu.cn; \{xuyunlu,~chengzhanzhan,~pushiliang,~niuyi\}@hikvision.com; wufei@cs.zju.edu.cn
}

\begin{document}

\maketitle
\begin{abstract}
Arbitrary text appearance poses a great challenge in scene text recognition tasks. Existing works mostly handle with the problem in consideration of the shape distortion, including perspective distortions, line curvature or other style variations.
\textit{Rectification} (\ie\textit{spatial} transformers) as the preprocessing stage is one popular approach and extensively studied. However, \textit{chromatic} difficulties in complex scenes have not been paid much attention on. In this work, we introduce a new learnable geometric-unrelated rectification, Structure-Preserving Inner Offset Network (SPIN), which allows the color manipulation of source data within the network. This differentiable module can be inserted before any recognition architecture to ease the downstream tasks, giving neural networks the ability to actively transform input intensity rather than only the spatial rectification. It can also serve as a complementary module to known spatial transformations and work in both independent and collaborative ways with them.
Extensive experiments show the proposed transformation outperforms existing rectification networks and has comparable performance among the state-of-the-arts. 
\end{abstract}

\iffalse
\input{sections/1_intro.tex}
\input{sections/2_related.tex}
\input{sections/3_methods.tex}
\input{sections/4_exp.tex}
\input{sections/5_con.tex}
\fi

%%%%%%%%%%%%%%%%%%1 introduction%%%%%%%%%%%%%%%%%%%%%%%%%
\section{Introduction}
Optical Character Recognition (OCR) has long been an important task in many practical applications. 
%Optical Character Recognition (OCR) has attracted great attention due to its various practical applications. 
Recently, recognizing text in natural scene images, referred to as Scene Text Recognition (STR), has attracted great research interests 
%attention 
due to the diverse text appearances and even the extreme conditions. % in which these scenes are captured. 
Modern techniques in deep neural networks have been widely introduced \cite{CRNN,RARE,R2AM,ASTER,GRCNN,FAN,AON,SSFL,SSDAN,SRN}. 

Since existing methods are quite powerful for regular text recognition, reading irregular shaped text has become a challenging yet hot research topic for computer vision community. As spatial transformer network (STN) \cite{STN} was proposed, RARE \cite{RARE} integrated it as a mainstreaming preprocessing procedure in the STR framework. % before the regular encoder-decoder recognition stage. 
Notice that the \textit{rectification module} does not require any extra annotations yet is effective in an end-to-end network. Therefore, it was soon popularly utilized and further explored in later works \cite{AFDM,wrong,ASTER,ESIR,MORAN,ScRN}. 
\begin{figure}[!t]
	\centering
	\includegraphics[scale=0.45]{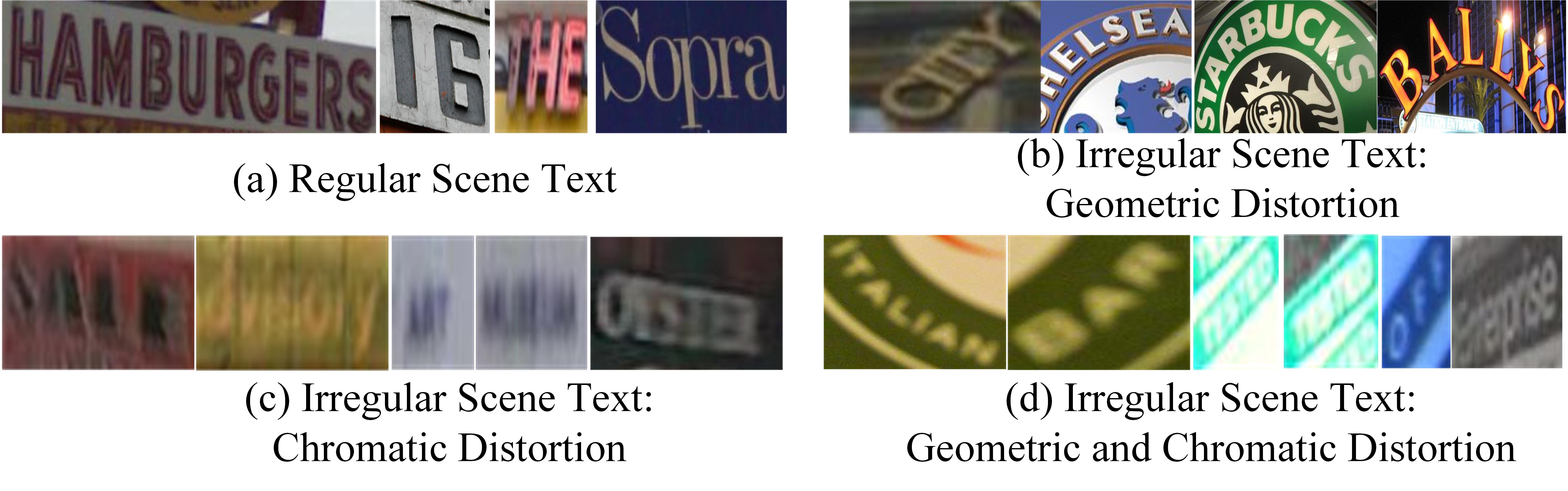}
	\caption{Examples of regular and irregular scene text, where the problems of irregular case can be categorized into geometric distortion and chromatic distortion.}
	\label{fig:0}
\end{figure}
Although almost all the existing transformations are limited to geometric rectification, shape distortions do not account for all difficult cases in STR. Severe conditions resulting from intensity variations, poor brightness, shadow, background and imaging noise, even some insensible to human beings also contribute to hard samples for deep networks. %occlusion, low resolution,
In particular, corresponding to geometric distortion, we term these irregularities as \textit{chromatic distortions}, which are very common and intractable but attracted rare attention in STR.
Early work \cite{landau1988importance} stated that humans prefer to categorize objects based on their shapes, even the geometric distorted texts in Fig. \ref{fig:0}(b) can be easily recognized on condition that \textit{the shapes of objects are well preserved}.
However, chromatic distortion leads to unclear shapes like in Fig.\ref{fig:0}(c) and (d), thus, recognition becomes much harder.
To reduce the burden of STR, we are inspired to rectify the chromatic distortion to restore shapes of texts.
That is, to design intensity manipulations towards variable distorted conditions.

Usually, chromatic distortions can be grouped into two cases (in Fig. 2) named \textit{inter-pattern} and \textit{intra-pattern} problems (explained later).
Here we denote all the pixels with the same intensity as a \textit{structure-pattern} (short in \textit{pattern}).
Assuming an ideal condition that the images to be identified are clearly composed of two distinct \textit{pattern}s: the intended text \textit{pattern} and the uncaring background \textit{pattern}, networks can focus on the intrinsic shape or features of the text \textit{pattern}.
However, real scenes are far more complex than the above assumption,
ubiquitous images are facing diverse chromatic challenges.
{
In the detail of two chromatic distortion types,
(a) \textit{inter-pattern problem} means noise \textit{pattern}s are close to the text \textit{pattern}s (\eg poor contrast or brightness) or intensities of close text \textit{pattern}s are dispersed. 
%Thus, 
%it needs re-balance between \textit{pattern}s (\ie separating the different objects and aggregating homogeneous ones) while maintaining the consistency within each \textit{pattern}. As in Fig.2 (a), text becomes easier to recognize after intensities of `L',`O',`V',`E' and background are pulled away, while the intensities of the characters keep close.
Thus, 
it needs separate the text \textit{patterns} and background \textit{patterns}, and meanwhile aggregate characters into the uniform text \textit{patterns}. % while maintaining the consistency within each \textit{pattern}. 
As in Fig.2 (a), text becomes easier to recognize after intensities of `L',`O',`V',`E' and background are pulled away, while the intensities of the characters keep close.
(b) \textit{Intra-pattern problem} means text \textit{pattern}s are interfered by the noise like shade, occlusion \etc. As in Fig.2 (b), the shade on left-bottom is mixed with `L', which should be obviated.
}
Therefore, a good chromatic rectifier should have the ability to handle with both. % the above two 

\begin{figure}[!t]
	\centering
	\includegraphics[scale=0.58]{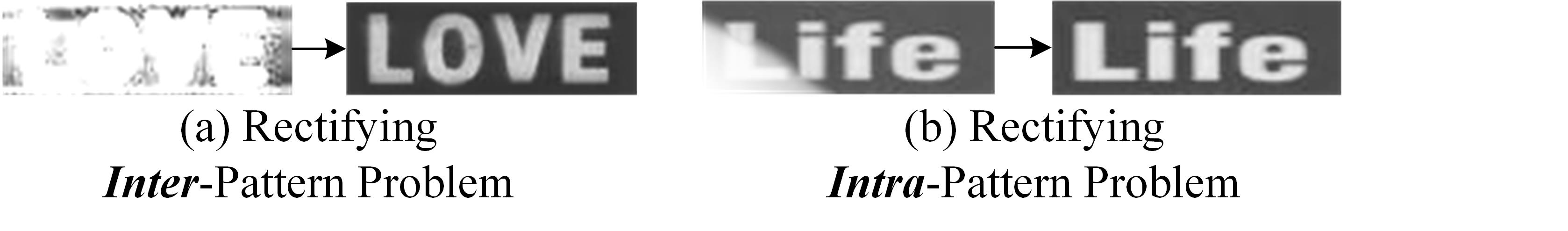}
	%\vspace{-0.6cm}
	\caption{Two types of chromatic difficulties. \textit{Pattern} is defined as all the pixels of the same intensity in an image.}
	\label{fig:0-1}
%\vspace{-0.6cm}
\end{figure}

In this work,
we propose a new chromatic rectifier called \textbf{\textit{SPIN}}, which stands for {\textit{\textbf{S}tructure-\textbf{P}reserving \textbf{I}nner Offset \textbf{N}etwork}}. Here, we borrow the concept of \emph{offset} from \cite{MORAN} intended for pixel-level spatial drift at birth, but we enrich it to a broader prospect, containing channel intensity offset (denoted in \textbf{inner offset}) and spatial offset (denoted in \textbf{outer offset}), respectively. The \textit{inner offset} is designed to mitigate chromatic distortions, such as the mentioned \textit{inter-} or \textit{intra-pattern} difficulties, while the \textit{outer offset} incorporates the geometric rectifications such as \cite{RARE,ASTER,MORAN,ESIR,ScRN}. 
Note that it is the first to focus on the \textit{inner offset} of scene text,
as the proposed SPIN mitigates chromatic distortions for a clearer shape of object.
{
Specifically, it consists of two components: the \textit{Structure Preserving Network (SPN)} and the \textit{Auxiliary Inner-offset Network (AIN)}. SPN is responsible for alleviating the irregularities caused by \textit{inter-pattern} problems. And AIN is an accessory network to distinguish the irregularities aroused by \textit{intra-pattern} problems.
}
These two components can complement each other.
The rectified images will thus become visually clearer for deep networks in intrinsic shape of texts and then easier to recognize. Moreover, we explore the integration of both the inner and outer offsets by absorbing geometric rectification (\ie TPS \cite{ASTER,wrong}) into SPIN. Experiments verify that, the unified inner and outer offsets in a single transformation module will assist each other leading to even better recognition.
The contributions can be summarized as:

(1) To the best of our knowledge, it is the first work to handle with the \textit{chromatic distortions} in STR tasks, rather than the extensively discussed spatial ones. We also introduce the novel concept of \textit{inner} and \textit{outer} offsets in rectification networks and propose a novel SPIN to rectify the images with chromatic transformation.

(2) The proposed SPIN can be easily integrated into neural networks and trained in an end-to-end way without additional annotations and extra losses. 
%Unlike the typical spatial transformation based on STN, which is tied to tedious initialization schemes \cite{RARE,ESIR} or stage-wise training procedures \cite{ScRN}, SPIN requires \textbf{no} need of sophisticated initialization or any network tricks, which enables it to be a more flexible module.
Unlike STN relying on tedious initialization schemes \cite{RARE,ESIR} or stage-wise training procedures \cite{ScRN}, SPIN requires \textbf{no} need of sophisticated initialization or any network tricks, enabling it to be a more flexible module.

(3) The proposed SPIN achieves impressive effectiveness in multiple STR tasks. And the combination of chromatic and geometric transformations has been experimentally proved to be practicable, which further outperforms existing techniques by a large margin.
%%%%%%%%%%%%%%%%%%%%%%%%%%%%%%%2 related%%%%%%%%%%%%%%%%%%%%%%%%%%%%%%%%%%%%%%%%%%%%%%%%
\section{Related Work}
\textbf{{Scene Text Recognition}}. Text recognition in natural scene images is one of the most important challenges in computer vision and many methods have been proposed. Conventional methods were based on hand-crafted features including sliding window methods \cite{SVT,wang-word}, connected components \cite{realtime}, histogram of oriented gradients descriptors \cite{su2014accurate}, and strokelet generation \cite{strokelets} \etc. 
In recent years, deep neural network was dominating the area. 
As recurrent neural
networks (RNNs) \cite{LSTM,GRU} were introduced and combined with CNN-based methods\cite{CRNN}, more sequence-to-sequence models \cite{R2AM,FAN,2DCTC,lyu20192d,DAN} were proposed. The attention mechanism was applied to a stacked RNN on top of the recursive CNN \cite{R2AM} and some revisions on the attention \cite{FAN} appeared. Recently, works further advanced the encoder-decoder structure with 2-Dimensional(2-D) ones including 2-D CTC \cite{2DCTC}, 2-D attention \cite{2DATT,lyu20192d} and character-based segmentation before semantic learning \cite{DAN,SRN}.
As modern techniques are powerful to deal with regular texts in scenes, irregular text recognition is still posing a challenging task.
Existing methods include decoder on 2-D featuremaps \cite{2DATT,2DCTC,DAN,SRN}, character-level supervision and guidance \cite{YangX}, generator-based network \cite{Synth}, and rectifications \cite{RARE,ASTER,MORAN,ESIR,ScRN}.

\noindent\textbf{{Rectification Network}}.
Among above-mentioned approaches handling with \textit{irregularity} of scene texts, rectification-based approaches are popularly equipped before recognition \cite{RARE,ASTER,MORAN,ESIR,ScRN}. 
A distinct advantage is that they do not require any additional annotations or loss functions, which is flexible yet effective \cite{wrong,Survey}.
In this area, all the existing rectifications are limited to the geometric irregularity. Following Spatial Transformer Network (STN) \cite{STN}, RARE \cite{RARE,ASTER} first used thin-plate-spline \cite{TPS} transformation and regressed the fiducial transformation points on curved text. Similarly, ESIR \cite{ESIR} iteratively utilized a new line-fitting transformation to estimate the pose of text lines in scenes, and ScRN \cite{ScRN} improved the control point prediction in STN. MORAN \cite{MORAN} rectified multi-object using predicted offset maps. Different from the image-level rectification, STAR \cite{STAR} proposed a character-aware network by using an affine transformation network to rotate each detected characters into regular ones.
Here, we also work on designing rectification modules to ease downstream recognition. 
Differently from all the existing works, we point out that the \textit{chromatic rectification} is also important. Our proposed network aims to rectify the input patterns from a broader perspective, and also handle with the color or brightness difficulties in STR tasks.

%%%%%%%%%%%%%%%%%%%%%%%%%%%%%%%3 method%%%%%%%%%%%%%%%%%%%%%%%%%%%%%%%%%%%%%%%%%%%%%%%%
\section{Methods}
\subsection{Chromatic Transformation}
The proposed chromatic transformation \textit{SPIN} %, which stands for Structure-Preserving Inner Offset Network, 
consists of two components: the Structure Preserving Network (SPN) and the Auxiliary Inner-offset Network (AIN).

\subsubsection{Structure Preserving Network (SPN)}
Inspired by Structure Preserving Transformation (SPT) \cite{SPT}, which cheats modern classifiers by degenerating the visual quality of chromaticity, we find that this kind of distortion is usually fatal to the deep-learning-based classifiers. The potential ability of controlling the chromaticity enlightens us to shed light on an even broader application conditioned on proper adaptation. In particular, we find that SPT-based transformations could also rectify the images of color distortions by intensity manipulation, especially for examples like Fig. \ref{fig:0} (c) and (d).
Formally, given the input image ${x} \in \mathcal{I}$, let $ x' \in \mathcal{I'}$ denote the transformed image. A transformation $\mathcal{T}$ on $\mathcal{I}$ is defined as follows:
\begin{equation}
x'(i,j) = \mathcal{T}[x(i,j)],
\end{equation}
where $x(i,j)$ or $x'(i,j)$ is the intensity of the input or output image at the coordinate $(i,j)$. Specifically, the general form of SPT is defined to be a linear combination of multiple power functions:
\begin{equation}
x'= \mathcal{T}(x)  =  sigmoid( \sum_i \omega_i x^{\beta_i} ), 
\end{equation}
where $\beta_i$ is the exponent of the i-th basis power function, $\omega_i$ is the corresponding weight. 
{
With normalized the intensity of each pixel in [0,1], total intensity space can be scattered into K subspace. Each space can be modeled by an exponential function under certain linear constraint as $y_i = x_i^{\beta_i}$, where $\quad x_i =\frac{i}{2(K+1)}$, $y_i=1-\frac{i}{2(K+1)}$, i=1,2,3,...,K+1.
Then 2K+1 parameters in pair can be formulated as:
\begin{equation}
\beta_i  = \left\{
\begin{aligned}
& round(\frac{log(1-\frac{i}{2(K+1)} )}{log\frac{i}{2(K+1)}}, 2),  1\leq i  \leq K+1  \\
& round(\frac{1}{\beta_{i-(K+1)}},2),K+1 < i \leq 2K+1.
\end{aligned}
\right. 
\end{equation}
These $\beta$ exponents can be chosen based on domain or fixed in advance for simplicity.
{
K defines the complexity of transformation intensities. A larger K will support a more complex and fine-grained chromatic space.}
$\omega$ is generated from the input image with convolutional blocks. Therefore, the transformation based on SPN can be summarized as:
\begin{equation}
x'= \mathcal{T}(x)  =  sigmoid( \sum_i  W^s_{(i)}\mathcal{H_{SPN}}(x) x^{\beta_i} ),
\end{equation}
where $W^s$ and $\mathcal{H}_{SPN}(\cdot)$ are part of Block8 weights and the feature extractor, respectively.
$W^s\mathcal{H}_{SPN}(\cdot)$ is a \textit{(2K+1)} dimensional segment from Block8 output of totally \textit{(2K+2)} dimensions, as shown in Fig \ref{fig:1}(a). 
\begin{figure*}[!t]
	\centering
	\includegraphics[scale=0.98]{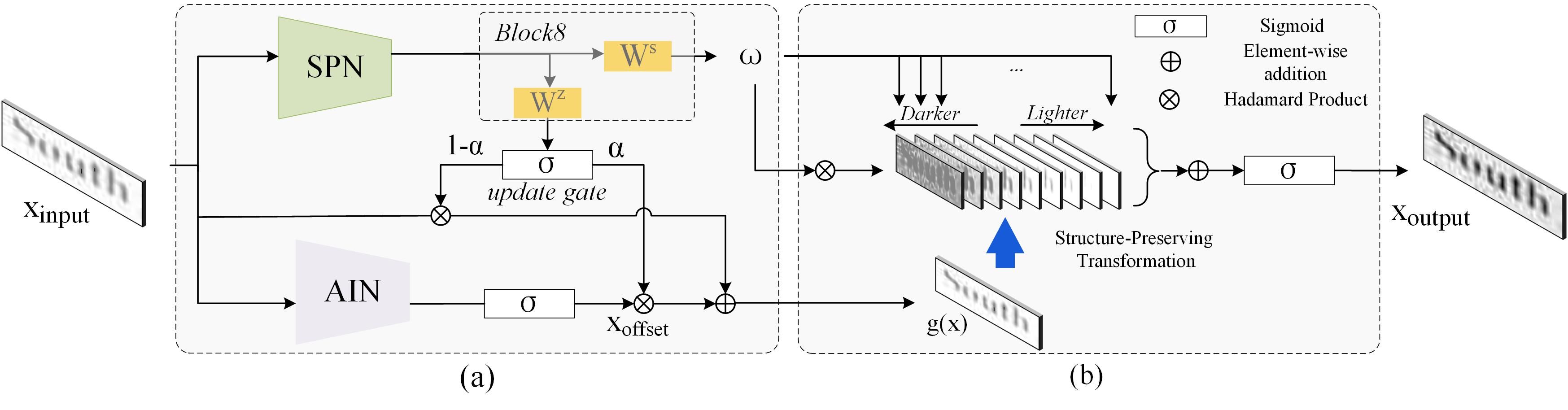}
	%\vspace{-0.3cm}
	\caption{The overview of SPIN. (a) The input image x would first be feed into a well-designed network, and output the updated image x and a group of weights respectively. (b) A structure-preserving transformation is performed on the updated image with the generated weights. } 
	\label{fig:1}
	%\vspace{-0.5cm}
\end{figure*}

\begin{table}[!h]
	\centering
	\scalebox{0.72}{\begin{tabular}{|c|c|c|c|c|}
			\hline
			\multirow{2}{*}{Layers} & \multicolumn{2}{c|}{Configurations}                 &
			\multicolumn{2}{c|}{Output} \\ \cline{2-5}
			& SPN                      & AIN                      & SPN                     &  AIN                        \\ \hline
			Block1                  & \multicolumn{2}{c|}{Conv(32, 3, 2, 2)}       & \multicolumn{2}{c|}{50$\times$16}                   \\ \hline
			Block2                  & \multicolumn{2}{c|}{Conv(64, 3, 2, 2)}       & \multicolumn{2}{c|}{25$\times$8}                    \\ \hline
			Block3                  & \multicolumn{2}{c|}{Conv(128, 3, 2, 2)}      & \multicolumn{2}{c|}{12$\times$4}                     \\ \hline
			Block4-1/Block4-2       & Conv(256, 3, 2, 2) & Conv(16, 3, 2, 1) & 6$\times$2&6$\times$2                \\ \hline
			Block5-1/Block5-2       & Conv(256, 3, 2, 2)& Conv(1, 3, 1, 1)  & 3$\times$1&6$\times$2                \\ \hline
			Block6                  & Conv(512, 3, 1, 1)  & -                        & 3$\times$1& -                     \\ \hline
			Block7                  & Linear(256)          & -                        & 256 & -                    \\ \hline
			Block8                  & Linear(2K+2)          & -                        & 2K+2 &    -                 \\ \hline
	\end{tabular} }
	\label{tab:1}
    %\vspace{-0.1cm}
	\caption{The architecture of SPN and AIN. The parameters in Conv($\cdot$) are filter numbers, kernel, stride and padding size of convolution orderly. For linear blocks, the value in Linear($\cdot$) means output channel numbers. Weights from Block1 to Block3 are sharing.}
	%\vspace{-0.2cm}
\end{table}
In essence, \textbf{Structure-Preserving} is realized by filtering the intensity level of input images.
All the pixels with the same intensity level in the original image have the same intensity level in the transformed image, where the set $\{(i,j)|x(i,j)=c\}$ with intensity level c is defined as \textit{structure-pattern}s. Intuitively, we propose the SPN for rectifying chromatic distortions by taking advantage of this singleton-based pixel-wise transformation in \textit{two aspects}:  (1) Separating useful structure-patterns from the harmful ones by injecting them to different intensity levels, which will be likely to generate better contrast and brightness. (2) Aggregating different levels of structure-patterns by mapping them to close intensity levels, which will be beneficial to alleviate fragments, rendering a more unified image. These are suitable for handling with \textit{inter-pattern} problems (as Fig.2 (a)) but powerless against the other.

\subsubsection{Auxiliary Inner-offset Network (AIN)}
\label{ION}
Since SPN tries to separate and aggregate the specific \textit{structure-pattern}s by exploiting the spatial invariance of words or characters, it inexplicably assumes that these patterns are under inconsistent intensities, namely different levels of \textit{structure-pattern}s. However, it does not consider the disturbing patterns can have similar intensity with the useful ones, noted as \textit{pattern confusion}, causing \textit{intra-pattern} problems (as Fig.2 (b)). 

To handle with the difficulty, we borrow the concept of \emph{offset} from geometric transformation \cite{MORAN} while we decouple geometric and chromatic offsets for better understanding of the rectification procedures. SPN will generate \textbf{chromatic offsets}(namely \textit{inner-offsets}) on each coordinate. As mentioned above, solely SPN is limited by the intrinsic \textit{pattern confusion} problems. Here, we design an auxiliary inner-offset network (AIN) to assist with the consistent intensity for patterns. The \textit{auxiliary inner-offset} is defined as:
\begin{equation}
g(x) = (1-\alpha) \circ x + \alpha \circ x_{\text{offsets}}, 
\end{equation}
where
\begin{equation}
\alpha = sigmoid(W^z \mathcal{H_{SPN}}(x)),
\end{equation}
\begin{equation}
x_{\text{offsets}} = W^a \mathcal{H_{AIN}}(x). 
\end{equation}
$W^z$ is the trainable scalar which is partial output from Block8 in Fig \ref{fig:1}(a). Similarly, $W^a$ is the trainable parameter in AIN module and $\mathcal{H_{AIN}}(\cdot)$ is feature extractor sharing the first 3 blocks with SPN. $\circ$ is hadamard product. We design a learnable update gate $\alpha$, which can receive signals from SPN and perceive the difficulty of different tasks. It is responsible for controlling the balance between the input image x and the predicted auxiliary inner-offsets. $g(x)$ (or $x$) is the updated (or input) image. Here, $x_{\text{offsets}}$ is predicted by AIN, whose architecture is also given in Table 1.
As chromatic transformation is pixel-to-pixel mapping on each coordinate and thus requires no need of spatial shift. The AIN first divides the image into small patches and then predicts offsets for each. All the offset values are activated by $sigmoid(\cdot)$ and mapped into the size of input image via common upsampling (\eg{bilinear interpolation}). 
The auxiliary inner-offsets can mitigate the \textit{pattern confusion} through slight intensity-level perturbations $x_{\text{offsets}}(i,j)$ on each coordinate $(i,j)$.
Assisted by AIN, the enhanced transformation will be performed on the updated images, and the comprehensive transformation $\hat{\mathcal{T}}$ can be formulated upgrading Equa. (2) by
\begin{equation}
\hat{x'} =  \hat{\mathcal{T}}(x)  =  sigmoid( \sum_i \omega_i (g(x))^{\beta_i} ).
\end{equation}
The detailed network and the overall structure are also illustrated in Table 1 and Fig. \ref{fig:1}.

\subsection{Geometric-Absorbed Extension}
Spatial transformations \cite{STAR,RARE,ASTER,ESIR,MORAN} rectify the location shift of patterns by predicting the corresponding coordinates, which will generate geometric offsets (namely \textit{outer-offset}s). And then they re-sample the whole images based on the points, which could be described as:
\begin{equation}
\tilde{x'}(i',j')= \mathcal{S}(x, f(i', j'))=\mathcal{S}(x, (i, j)),
\end{equation}
where $(i,j) (or (i',j'))$ is the origin coordinate (or the coordinate adjusted by outer-offsets), $\mathcal{S}(x,\cdot)$ here stands for the sampler, which generates the transformed images by interpolating the neighbor pixels in $x(i,j)$, and $f$ represents different forms of  transformation function, \eg Affine \cite{STAR,RARE}, Thin-Plate-Spline(TPS) \cite{TPS,RARE,ASTER,ScRN}, Line-Fitting Transformation \cite{ESIR}.
Differently, in the proposed chromatic transformation generate \textit{inner-offsets} on each coordinate as Equa. (1). Both are independent modules used to ease downstream stages, while they can be integrated as the unified transformation as
\begin{equation}
\begin{aligned}
\ddot{x'}(i',j')&= \mathcal{S}(\hat{x'}, f(i', j')) \\
&= \mathcal{S}(\hat{\mathcal{T}}(x),  f(i', j')).
\end{aligned}
\end{equation}
We call the uniform of chromatic and geometric rectification as Geometric-Absorbed SPIN (short in \textit{GA-SPIN}).
Note that the learnable parameters in both chromatic transformation $\hat{\mathcal{T}}(\cdot)$ and geometric transformations $f(\cdot)$ are simultaneously predicted by the unified network configurations in Table 1. When absorbing spatial rectifications like TPS, the only difference for GA-SPIN compared to SPIN is the output channel numbers of Block8 are set to $2K+2+N$ instead of $2K+2$, as N is number of parameters for TPS (\eg 40).

\subsection{Recognition Network and Training}
For direct comparison on rectification, we adopt the mainstreaming recognition network \cite{wrong} and all its configuration and setups. 
Specifically,
for contextual feature, we use typical STR backbones of 7-layer of VGG \cite{VGG} or 32-layer ResNet \cite{ResNet} with two layers of Bidirectional long short-term memory (BiLSTM) \cite{LSTM} each of which has 256 hidden units as \cite{CRNN,FAN}.
Following \cite{wrong}, the decoder adopts an attentional LSTM with 256 hidden units, 256 attention units and 69 output units (26 letters, 10 digits, 32 ASCII punctuation marks and 1 EOS symbol).

%%%%%%%%%%%%%%%%%%%%%%%%%%%%%%%4 exp%%%%%%%%%%%%%%%%%%%%%%%%%%%%%%%%%%%%%%%%%%%%%%%%
\section{Experiments}
\begin{table*}[!htb]
	\centering
	\scalebox{0.88}{
		\begin{tabular}{|p{1.5cm}<{\centering}|l|*{2}{p{0.7cm}<{\centering}}|*{7}{p{0.75cm}<{\centering}}|p{0.8cm}<{\centering}|}
			\hline
			\multicolumn{1}{|c|}{\multirow{2}{*}{Rectification}} &
			\multicolumn{1}{c|}{\multirow{2}{*}{Transformation}} & \multicolumn{2}{c|}{Offsets}        & \multicolumn{8}{c|}{Benchmark}     \\ \cline{3-12}
			\multicolumn{1}{|c|}{}   &  \multicolumn{1}{c|}{}                           & \multicolumn{1}{c}{Outer} & Inner & IIIT                 & SVT                  & IC03                 & IC13                 & IC15                 & SVTP                 & CT                  & \multicolumn{1}{c|}{ Avg}         \\ \hline
			None &(a) \footnotesize{None$^\dagger$ \cite{RARE}  }                                             & No                         & No    &      82.4                &     81.9                 &   91.2                   &   87.3                   &   67.4                   &  70.9                    &   61.6                    &   77.5            \\ \hline
			\multirow{2}{*}{Chromatic} &(b) \footnotesize{SPIN w/o AIN }                                            & No                         & Yes    &
			82.1                   &   84.2                  &   91.9                &  87.8                   &        67.3                   &   72.3                   &  63.5                 &   78.4                   \\
			%(c) \footnotesize{+ Scheme Initialization}                                            & No                         & Yes   &
			%82.5                    &   82.8                &  91.2                   &   87.9                   &  67.3                    &   74.1                   &   63.9                    &   78.5             \\
			&(c) \footnotesize{SPIN}                                           & No                         & Yes   &   82.3                   &   84.1                  &  91.6                    &    88.7                  &   68.1                   &    72.3                  &   64.6                    &    78.8                 \\\hline
			Geometric &(d) \footnotesize{STN$^\dagger$ \cite{wrong}}                                      & Yes                        & No    & \multicolumn{1}{c}{82.7} & \multicolumn{1}{c}{83.3} & \multicolumn{1}{c}{92.0} & \multicolumn{1}{c}{88.5} & \multicolumn{1}{c}{69.6} & \multicolumn{1}{c}{74.1} & \multicolumn{1}{c|}{62.9} & \multicolumn{1}{c|}{79.0}  \\\hline
			\multirow{2}{*}{Both} &(e) \footnotesize{SPIN+STN }                                            & Yes                        & Yes   &  83.6                    &   84.4                   &   92.7                   &  89.2                    &  70.9                    &   73.2                   &   64.6                    &  79.8                     \\ %\hline
			&(f) \footnotesize{GA-SPIN}     & Yes & Yes & 84.1 &83.2 &92.3&88.7&71.0&74.5&67.1&80.1 \\\hline
		\end{tabular}
	}
    \caption{Comparison between the baseline and the model combined with different rectifications. All the models are evaluated based on accuracy under several benchmarks. The `Offset' stands for enabling the corresponding offset modules. `$\dagger$' indicates the model is tested by us under a fair setting.} %}
	\label{table:2}
\end{table*}

\subsection{Dataset}
Models are trained only on 2 public synthetic datasets MJSynth (MJ) \cite{MJ} and SynthText (ST) \cite{ST} without any additional dataset or data augmentation.
We evaluate on 4 regular and 3 irregular datasets as follows according to the difficulty and geometric layout of the text \cite{wrong}.

\noindent \textbf{IIIT5K (IIIT)} \cite{IIIT5K} contains scene texts and born-digital images. It consists of 3,000 images for evaluation.

\noindent \textbf{SVT} \cite{SVT} contains 647 images for evaluation, collected from Google Street View, some of which are noisy, blurry, or of low-resolution.

\noindent \textbf{ICDAR2003 (IC03)} \cite{IC03} contains 860 images for evaluation which was used in the Robust Reading Competition in the International Conference on Document Analysis and Recognition (ICDAR) 2003.

\noindent \textbf{ICDAR2013 (IC13)} \cite{IC13} contains 1015 for evaluation which was used in the ICDAR 2013.

\noindent \textbf{ICDAR2015 (IC15)} \cite{IC15} contains 2,077 text image patches are cropped scene texts suffer from perspective and curvature distortions, which was used in the ICDAR 2015. Researchers have used two different versions for evaluation: 1,811 and 2,077 images. 
We use 1,811 images in our discussions and report both results when compared with other techniques for fair comparisons.

\noindent \textbf{SVTP} \cite{SVTP} contains 645 images for evaluation. Many of the images contain perspective projections due to the prevalence of non-frontal viewpoints.

\noindent \textbf{CUTE80 (CT)} \cite{CUTE80} has 288 cropped images for evaluation. Many of these are curved text.

\subsection{Implementation}
All images are resized to 32 $\times$ 100 before entering the network. $K=6$ is the default setting. The parameters are randomly initialized using He \etal's method \cite{He} if not specified. Models are trained with the AdaDelta \cite{adadelta} optimizer for 5 epochs with batch size = 64.
The learning rate is set to 1.0 initially and decayed to 0.1 and 0.01 at 4-th and 5-th epoch, respectively.

\textbf{NOTE} that the inconsistent choices of datasets, network and training configurations make fair comparison and module-wise assessment of recent works more difficult. We evaluate the proposed rectification module with SOTAs rigorously using the same datasets, recognition network and training setups as the unified benchmark \cite{wrong}. All experiments are only with word-level annotations. The network is trained from the scratch end-to-end without stage-wise procedures or other tricks.
\subsection{Effects of Rectification Module}
To go deep into individual factors in the proposed model, we cumulatively enable each configuration \textit{one-by-one} on top of a solid baseline. All models are trained on the MJ dataset with backbone of 7-layer VGG. Following models are all the same by default. Table \ref{table:2} lists the accuracy in each configuration.
\subsubsection{Chromatic Transformation.}
The item (a) poses the baseline \cite{RARE} without the rectification module. (b) and (c) add the rectification module and verify the advantage of the chromatic rectification. Specifically, (b) enables the proposed SPIN w/o AIN and improves (a) by an average of 0.9\%. And (c) moves forward to enable AIN with SPN, composing SPIN. Though only two additional convolution layers appended to (b), (c) obtains an average of 0.4\% consistent improvement and a total of 1.2\% improvement compared with the baseline. The enhancement is mainly from SVT, IC15 and CUTE80, where lots of images with chromatic distortions are included. In addition, the first 3 rows in Fig. 5, mostly the \textit{inter-pattern} distortions are transformed to clearer shape through SPIN-based method, and the last row, seemingly the \textit{intra-pattern} problems(\ie shadow), also show the obvious effect between SPIN over w/o AIN.
\subsubsection{Chromatic versus Geometric Transformation.}
Table 2(d) shows the baseline (a) equipped with STN \cite{wrong}, which rectifies images via \textit{outer-offset}s and achieved superior performance among the state-of-the-arts. Interestingly, the performance of SPIN in (e) is close to that of the geometric one. The difference lies in that, the former mainly improves the SVT and CUTE80 benchmarks, where large proportion of chromatically distorted images are included, while the latter mainly improves the IC15 and SVTP benchmarks, where rotated, perspective-shifted and curved images dominate. And it verifies the color rectification is comparably important with shape transformation and the two types of rectifiers possess their respective strengths.
\subsubsection{Combining Chromatic and Geometric Transformation.}
%We then explore the combination of both geometric and chromatic rectifications.
We further explore whether inner-offsets assist the existing outer-offsets \cite{RARE} and how these two categories can jointly work better.
In (e), simply placing the SPIN (c) before the TPS-based STN (d) (denoted as \textit{SPIN+STN}) without additional modification brings an average of 0.8\% improvement, compared with configuration (d).
It indicates that these two transformations are likely to be two complementary strategies, where the inner part was usually ignored by the former researchers.
GA-SPIN in (f) unifies SPIN and TPS-based STN in a single transformer (illustrated in Section 3.2).
In detail, the inner and outer offsets are predicted by two single networks in (e) and the total parameters are about 4M. While for (f), the pipelines are optimized using the unified network and the total parameters are 2.31M.
Though cutting parameters and computation cost compared to the straightforward pipeline structure (e), it improves the results by 0.3\% on average and gains a large margin by 1.1\% compared to solely geometric transformers (d). It means GA-SPIN is a simple yet effective integration. 
\subsection{Discussion}
\subsubsection{Discussion on Sensitivity to Weight Initialization.} Proper weight initialization is necessary for geometric transformations \cite{RARE,ASTER,ESIR}, since highly geometrically distorted images will ruin the training of the recognition network.
On the contrary, chromatic transformations are less likely to severely distort the images for the intrinsic merits of preserving the \textit{structure-pattern}s. Random initialization works well in Table \ref{table:2} (b) and (c), indicating insensitivity to weight initialization of our modules.
\begin{figure}[!htb]
	\centering
	\includegraphics[scale=0.33]{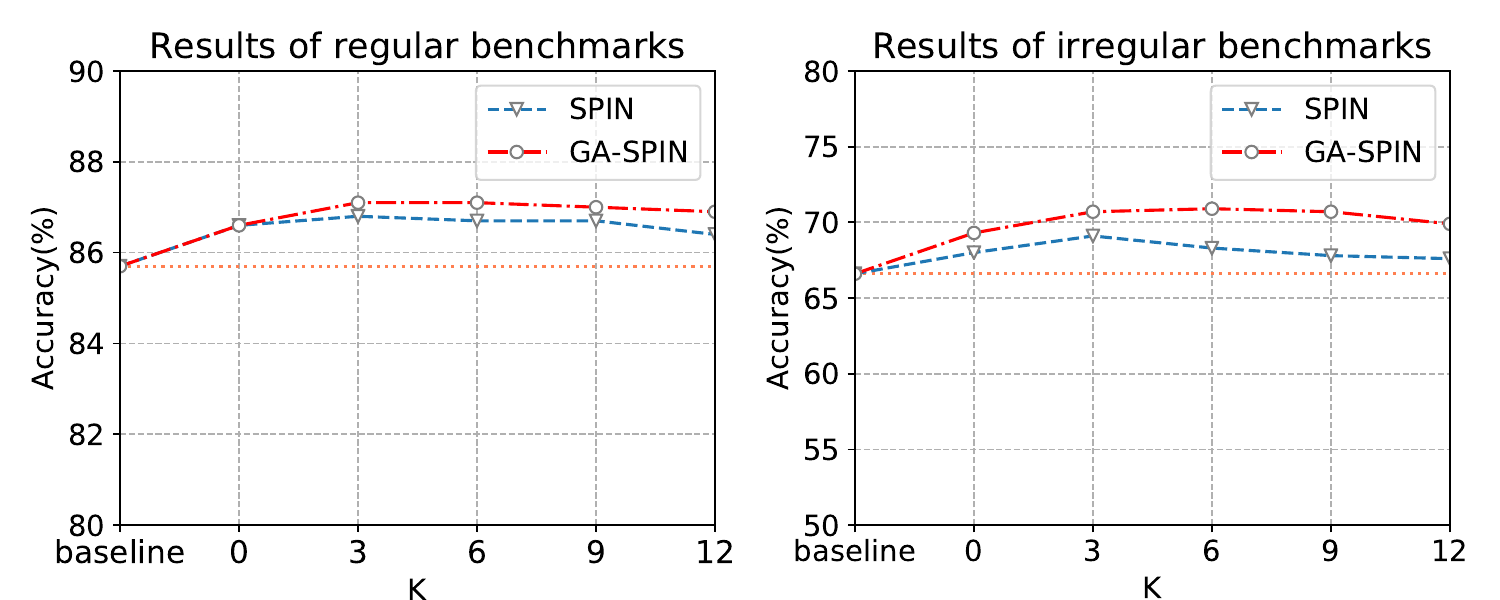}
	\caption{Results on different K. All the models are trained on the MJ dataset using VGG feature extractor. The line in coral, blue and red refer to the baseline without rectification module, with SPIN and with GA-SPIN, respectively. The results are evaluated with the mean accuracy.}
	\label{fig:3}
\end{figure}

\begin{table}[!b]
	\centering
	\scalebox{0.7}{
		\begin{tabular}{|c|c|c|c|c|c|} 
        \hline
		\multirow{2}{*}{Transformation} & \multicolumn{3}{c|}{Accuracy(\%)}& \multirowcell{2}{$\#$Para.\\($\times 10^6$)}&\multirowcell{2}{$\#$FLOPs\\($\times 10^{-9}$)}\\ \cline{2-4}
        &VGG-MJ & ResNet-MJ& ResNet & \\ \hline
		None     & 66.6          &  69.0  & 74.9 &0 &-\\
		STN \shortcite{wrong}$^\dagger$   & 68.9(+2.3)   &  71.2(+2.2)&76.9(+2.0) & 1.68&- \\
		
		ASTER \shortcite{ASTER}            & 68.2(+1.6) &  70.6(+1.6)$^*$ &78.0(+3.1)$^*$ &  - &0.12\\
		ESIR \shortcite{ESIR}                  & 70.2(+3.6)   &  72.7(+3.7)$^*$ &79.9(+5.0)$^*$ &  -  & -\\
        ScRN \shortcite{ScRN}                  & -    & \textbf{73.0(+4.0)}$^*$ & 82.3(+7.4)$^*$ & 2.62 &0.54\\
		\hline

		%SPIN w/o AIN                          & 67.7(+0.9)   & 69.9(+0.9) &- &2.28 \\
		SPIN                           & 68.3(+1.7)   &  71.0(+2.0)  &81.3(+6.4) &2.30&0.11\\
	    GA-SPIN               & \textbf{70.9(+4.3)} & \textbf{73.0(+4.0)} & \textbf{84.5(+9.6)}& 2.31& 0.11\\ \hline
	\end{tabular}
	}
	\caption{Comparison with rectification methods. The values are mean accuracy of 3 irregular benchmarks. `-MJ' means only using MJ dataset. `*' denotes the results with deeper 45/50-layer than our 32-layer ResNet setting. `$\dagger$' indicates the model is tested by us under the same setting for fairness.}
\label{table:2.1}
\end{table}

\begin{figure}[!t]
	\centering
	\includegraphics[scale=0.5]{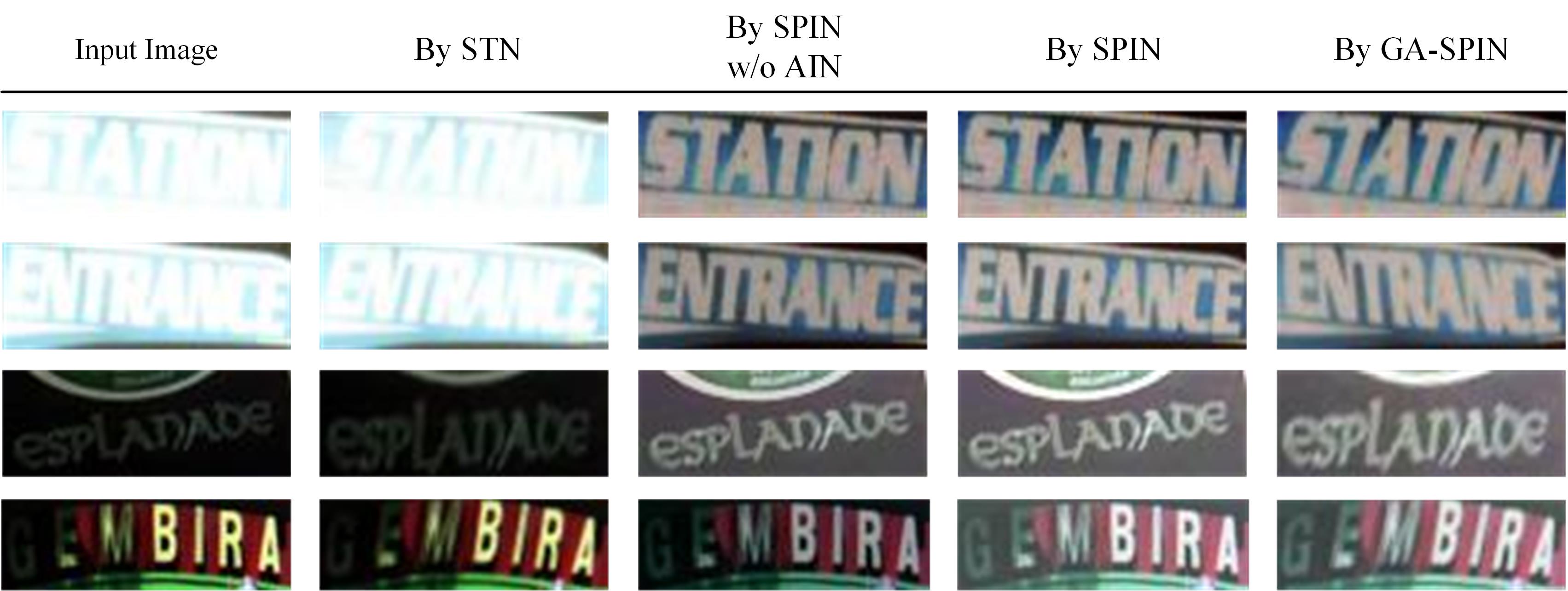}
    \caption{Visualization of rectification with chromatic problems, unbalanced contrast (row 1-2), low brightness (row 3) and shadow (row 4),respectively. Samples here and in Fig. \ref{fig:5} are all from SVT, IC15 and CUTE80 test sets.}
	\label{fig:4}
\end{figure}
\begin{figure}[!t]
	\centering
	\includegraphics[scale=0.6]{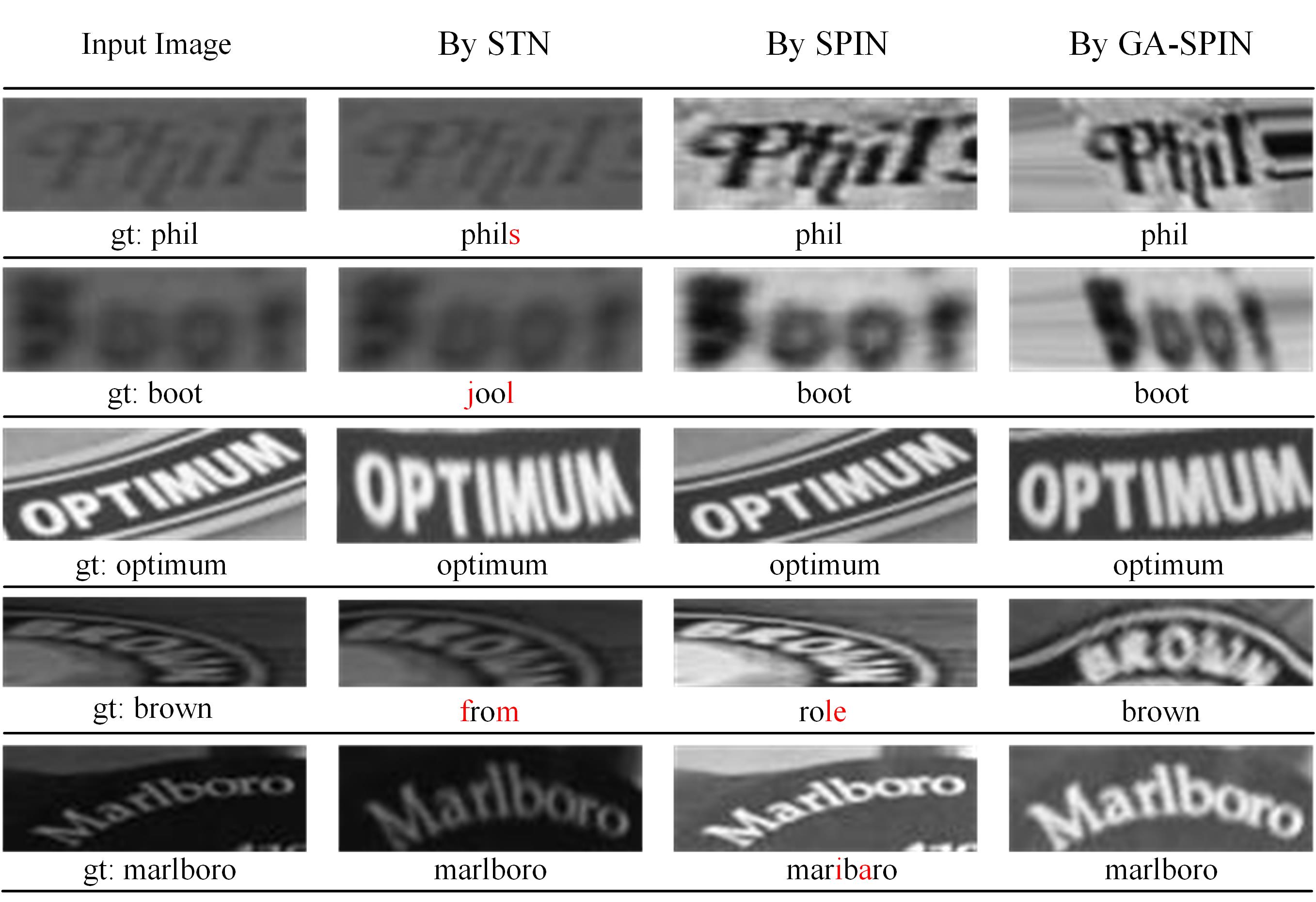}
	\caption{Examples of the rectification effects and prediction. ``gt'' is the ground truth. Predictions are shown under the images. Red are mistakenly recognized characters. }
	\label{fig:5}
\end{figure}

\begin{table*}[!t]
	\centering
	\scalebox{0.75}{
		\begin{tabular}{|l|l|l|l|l|*{8}{p{0.8cm}<{\centering}}|}
			\hline
			\multirow{2}{*}{Methods}  & \multirow{2}{*}{Rect}& \multirow{2}{*}{ConvNet} & \multirow{2}{*}{Training data} & \multirow{2}{*}{Supervision} & \multicolumn{8}{c|}{Benchmark}                                                                                                                                                  \\ \cline{6-13}
			&                          &    &              &              & IIIT                  & SVT                      & IC03                     & IC13                     & IC15a & IC15b                  & SVTP                  & CT                   \\ \hline
			Jaderberg \etal \cite{JaderbergReading}         &      & VGG                      & MJ            &   Word       & \multicolumn{1}{c}{-} & \multicolumn{1}{c}{80.7} & \multicolumn{1}{c}{93.1} & \multicolumn{1}{c}{90.8} & \multicolumn{1}{c}{-} & \multicolumn{1}{c}{-} & \multicolumn{1}{c}{-} & \multicolumn{1}{c|}{-} \\
			Shi \etal \cite{CRNN}             &        & VGG                      & MJ          &   Word                   & 78.2                  & 80.8                     & 89.4                     & 86.7                     & \multicolumn{1}{c}{-} & \multicolumn{1}{c}{-} & \multicolumn{1}{c}{-} & \multicolumn{1}{c|}{-} \\
			Shi \etal \cite{RARE}      &$\surd$              & VGG                      & MJ          &      Word              & 81.9                  & 81.9                     & 90.1                     & 88.6                     & \multicolumn{1}{c}{-} &    \multicolumn{1}{c}{-} & 71.8                  & 59.2                   \\
			Lee \etal  \cite{R2AM}          &        & VGG                      & MJ             &   Word                & 78.4                  & 80.7                     & 88.7                     & 90.0                     & \multicolumn{1}{c}{-} & \multicolumn{1}{c}{-} & \multicolumn{1}{c}{-} &  \multicolumn{1}{c|}{-} \\
			Shi \etal \cite{ASTER}        &$\surd$            & ResNet                   & MJ+ST        &   Word                  & {93.4}                  & 89.5                     &  94.5 & 91.8                    & 76.1                  & \multicolumn{1}{c}{-} & 78.5                  & 79.5                   \\
			
			Zhan \etal \cite{ESIR}     &$\surd$                & ResNet                   & MJ+ST       &     Word                 & 93.3                  & 90.2                   &  \multicolumn{1}{c}{-}                        & 91.3                     & 76.9                 & - & 79.6                & 83.3                  \\
			
			Baek \etal \cite{wrong} &$\surd$ & ResNet           & MJ+ST        &    Word             & 87.9  &
			87.5                       &{94.9}              &       92.3                & 77.6    & \multicolumn{1}{c}{71.8} & 79.2 &  74.0 \\
			
			Luo \etal \cite{MORAN}   &$\surd$                & ResNet                   & MJ+ST   &  Word      &91.2 &88.3       &    \textbf{95.0}                   & 92.4                           & -               & \multicolumn{1}{c}{68.8} &76.1 & 77.4                \\
			
			Xie \etal \cite{ACE}        &             & ResNet                   & MJ+ST  &   Word      &\multicolumn{1}{c}{-} &\multicolumn{1}{c}{-}      &    \multicolumn{1}{c}{-}                  & \multicolumn{1}{c}{-}                &-&    \multicolumn{1}{c}{68.9}     & 70.1               & 82.6                \\

			Wang \etal \cite{DAN}         &            & ResNet                   & MJ+ST    &    Word                     &94.3                 & 89.2                   &  \textbf{95.0}                       & {93.9}              & -                  & \multicolumn{1}{c}{74.5} & 80.0                & 84.4                  \\
			Luo \etal \cite{Aug}          &           & ResNet                   & MJ+ST  &    Word                     &-                 & -                   &  -                      & -                     & -                  & 76.1  & 79.2                & 84.4                  \\
			Yue \etal \cite{RobustScanner}          &           & ResNet                   & MJ+ST  &    Word                     &\textbf{95.3}                 & 88.1                   &  -                      & \textbf{94.8}                     & -                  & 77.1  & 79.5                & \textbf{90.3}                  \\
			Zhang \etal \cite{AutoSTR}          &           & ResNet $\star$                 & MJ+ST &    Word                     &{94.7}                 & \textbf{90.9}                   &  93.3                      & {94.2}                     & {81.8}                  & -  & {81.7}                & -                  \\

			\hline
			Cheng \etal \cite{FAN}  & & ResNet               & MJ+ST       &    Word+Char          & 87.4                  & 85.9                     & 94.2                     &  93.3                        & 70.6                  &\multicolumn{1}{c}{-} &  \multicolumn{1}{c}{-}                  &   \multicolumn{1}{c|}{-}  \\
			Liao \etal \cite{Perspective}        &             & ResNet                   & MJ+ST    &    Word+Char                     &91.9                 & 86.4                   &  -                       & 91.5                     & -                  & - & -                & 79.9                  \\
			Yang \etal \cite{ScRN}        &$\surd$             & ResNet                   & MJ+ST  &    Word+Char                     &94.4                 & 88.9                   &  95.0                       & 93.9                     & -                  & {78.7} & 80.8                & 87.5                 \\
			Wan \etal \cite{TextScanner}     &                & ResNet                   & MJ+ST  &    Word+Char                     &93.9                 & 90.1            & -                   & 92.9                  & 79.4  & -                & 83.3 & 79.4                   \\
	
			Li \etal \cite{2DATT}        &             & ResNet                   & MJ+ST+P   &  Word      &91.5 &84.5       &    \multicolumn{1}{c}{-}                   &91.0                           & -                & \multicolumn{1}{c}{69.2} & 76.4 & 83.3                \\			
			Yu \etal \cite{SRN}          &           & ResNet                   & MJ+ST+P  &    Word                     & 94.8                 & 91.5                   &  -                 & \textit{95.5}                     & {82.7}                 & -  & {85.1}                & 87.8                  \\
			
			Hu \etal \cite{GTC}      &               & ResNet                   & MJ+ST+P  &    Word                     & \textit{95.5}                 & \textit{92.9}                   &  \textit{95.2}                  & 94.3                     & 82.5                 & -  & \textit{86.2}                & {92.3}                  \\		
            Yue \etal \cite{RobustScanner}          &           & ResNet                   & MJ+ST+P  &    Word                     &{95.4}                 & 89.3                   &  -                      & 94.1                     & -                  & 79.2  & 82.9                & \textit{92.4}                  \\
			\hline
			SPIN      &$\surd$         & ResNet                   &  MJ+ST         &  Word                  &   {94.7}                    & 87.6                         &   93.4                       &  91.5                        &  79.1                     & \multicolumn{1}{c}{76.0} & 79.7                     &  85.1                      \\
			GA-SPIN    &$\surd$             & ResNet                   & {MJ+ST}         &  Word                  &   95.2                    & \textbf{90.9}                         &  94.9                        & \textbf{94.8}                         &  \textbf{82.8}                     & \textbf{79.5} &  \textbf{83.2}                     &  {87.5}                      \\
			\hline
		\end{tabular}
	}
	\caption{Comparison with SOTA methods. `P' indicates additionally using extra synthetic or real datasets apart from MJ and ST. `Word' and `Char' in `Supervision' column means word-level and character-level annotations. For IC15, a and b indicate 1811 and 2077 examples, respectively. `Rect' indicates methods focused on rectifications. Method with `$\star$' indicates using multiple backbones for each test set while the others use only one for all test sets. For fair comparison, reported results using additional private data or annotations are not taken into account (Top result in \textit{italic}). Top accuracy for each benchmark is shown in \textbf{bold}. }
%Top accuracy without benchmark validation set is \underline{underlined}.`$\ast$' means using validation set from benchmarks.
	\label{table:3}
\end{table*}

\subsubsection{Discussion on Different Values of K.}
We empirically analyze the performance under different K values, from 0 to 12, 
demonstrated in Fig. \ref{fig:3} with mean accuracy under regular and irregular cases.
{
Both SPIN and GA-SPIN steadily improve the performance when K becomes larger to some extent.
We do not observe additional performance gain when K is unreasonably large as the smaller K indicates simpler transformation and the larger K may result in complexity and redundancy. While K=3 seems enough for regular text, larger (\eg K=6,9) is better for irregular benchmarks.}
\subsubsection{Discussion on Traditional Image Preprocessing.}
As chromatic difficulties are rarely seen in STR networks, we consider traditional image processing techniques for color correction (\eg equalization, binarization, morphological operations).
We try these techniques as preprocessing stage in our recognition task similarly in train and inference phase.
However, they do not work well (even not better than baseline). We attribute it to 2 main reasons:
(1) They suffer from signal loss through image processing operations before flowing into the deep network.
(2) They are manually designed for better image quality according to human intuition and visual feeling, not adapted to variable conditions. 
\subsubsection{Discussion on Promotion to Geometric Transformation.}
As mentioned, the union of the two can further boost the results (Table 2(e) or (f)). Table \ref{fig:supp5} further shows that GA-SPIN even has superior effects not only in chromatic but also in the geometric perspective over the single STN. We attribute it to the chromatic promotion for the geometric transformation, similar in the 4th row (the word `brown') in Fig. 6. It means the SPIN eases the burden for recognition networks through chromatic rectification, and thus helps the training of spatial transformation in GA-SPIN.
\begin{figure}[!ht]
	\centering
	\includegraphics[scale=0.58]{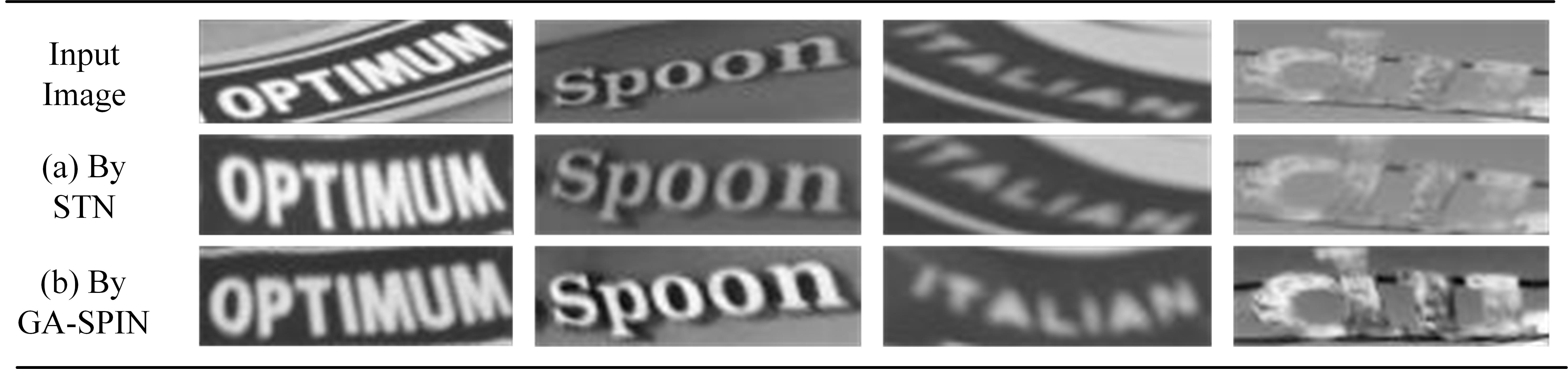}
	\caption{Examples of the rectification effects focused on the spatial distortions. Given input images with shape irregularities, (a) shows the results through STN-based transformation and (b) shows the results by the proposed GA-SPIN method. The downstream recognition networks after the rectification modules are the same for fair comparison.} 
	\label{fig:supp5}
\end{figure}
\subsubsection{Discussion on Training Procedures.}
Existing rectification modules are not easy to train as complained in \cite{ASTER,MORAN,ScRN}. The proposed SPIN requires no need of sophisticated initialization, or stage-wise training strategies.
Through the whole training procedures (visualized after 3 and 5 epoches in Fig. \ref{fig:supp3}),
we find that even after 3 epoches SPIN is almost completely trained, while the geometric rectification are still on the way to perform better. It shows that compared to geometric distortions, SPIN is much easier to learn and converge, which can be convenient to equip to variant networks.
\begin{figure*}[!t]
	\centering
	\includegraphics[scale=1.18]{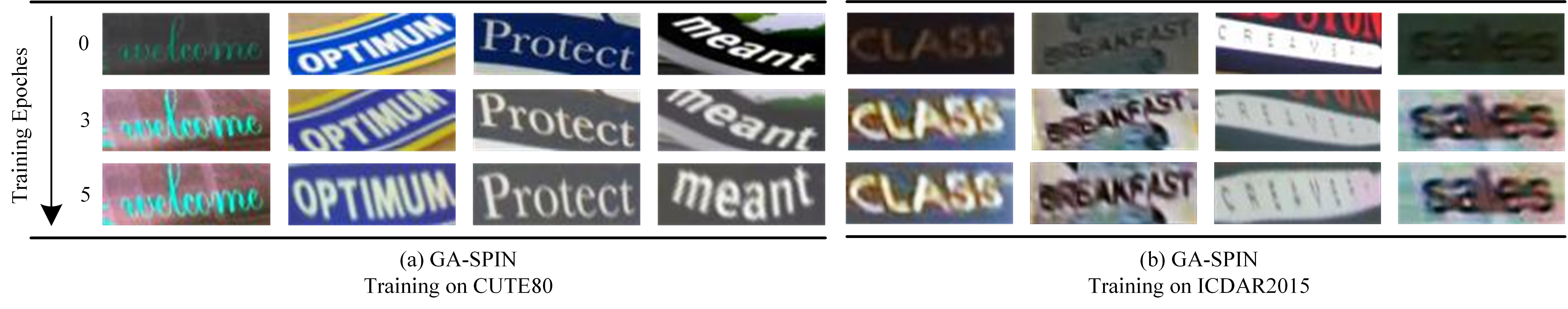}
 	\caption{Visualization of GA-SPIN training procedures. We show the input image and the rectified images after 3 and 5 epoches in each line, respectively.} %Samples are from SVT, IC15 and CUTE80.}
	\label{fig:supp3}
\end{figure*}
	
\subsection{Comparison with STN-based Methods}
We compare SPIN and GA-SPIN with existing rectification networks in both quantitative and qualitative evaluation. 
\subsubsection{Quantitative Results. }
Table \ref{table:2.1} lists the top results among existing geometric-based transformations \cite{ASTER,ESIR,ScRN} on irregular datasets with different backbones (VGG-7/ResNet-32) and different training set (MJ/MJ+ST). Both SPIN and GA-SPIN achieve superior recognition performance compared to existing methods. Note the recent ScRN\cite{ScRN} also has impressive performance, but its backbone chooses FPN with ResNet-50 and uses additional explicit supervision with char-level annotations of SynText \cite{ST}.

\subsubsection{Qualitative Results. }
Some visualization in Fig. \ref{fig:4} clearly shows the proposed SPIN can rectify the chromatic distortions by adjusting the colors.
The typical chromatic inter-pattern (row 1-3) and intra-pattern (row 4) problems which geometric rectifications do not consider, can be mitigated.

Fig. \ref{fig:5} illustrates the rectification effects and text predictions by STN \cite{ASTER}, SPIN and GA-SPIN, respectively. It shows rectifying chromatic distortion can improve the recognition results from two aspects: (1) It makes characters easier to recognize directly. The first 2 rows show that models tend to be misled by special symbols or similar shapes under severe chromatic distortions. At the same time, the proposed SPIN does not over-rectify or degrade images free of chromatic problems, such as the samples in the 3rd row. (2) Chromatic and geometric rectification can promote each other, as clearly shown in the last 2 rows of distorted images with both difficulties. Note in the 4th row, GA-SPIN has superior effects not only in chromatic but also in the geometric perspective over STN-based methods. We attribute it to the chromatic promotion for the geometric.

\subsection{Comparison with the State-of-the-Art}
Table \ref{table:3} lists the reported the state-of-the-art results on STR tasks. The proposed SPIN-based rectification network achieves impressive performance across the 7 datasets. 
\subsubsection{Rectification-based Approaches.}
Networks focused on rectification are all denoted in `$\surd$' in Table 4. Although with only word-level annotations, SPIN-based already outperforms all the methods using spatial transformers \cite{ASTER,wrong,ESIR,ScRN} on overall performance(especially in more challenging IC15, SVTP and CT datasets), even when recent method \cite{ScRN} applied heavier backbone and additional char-level annotations.
It points out that despite the widely focused geometric irregularity, color distortions are vital in these complex scenes and SPIN-based methods can effectively boost current performance.

\subsubsection{Overall Performance among SOTAs.}
Note that even with strong settings like character-level annotations \cite{FAN,Perspective} or deeper convolutional structures (\eg 45-layer ResNet \cite{ASTER,MORAN}) and newly designed advanced framework \cite{DAN}, even compared to reported results using additional real or private data (\eg \cite{TextScanner, ScRN}), SPIN obtains satisfying performance and GA-SPIN shows significant promotion.
Our methods only slightly fall behind (on solely IC03 dataset) the best reported \cite{DAN,MORAN} by 0.1\%, and we attribute the reason to stronger feature extractor (\ie deeper convolution and multi-scale features) and advanced decoupled attention mechanism \cite{DAN}. Advanced networks bring strong ability on regular recognition but fall behind GA-SPIN by a large margin on more complex scenes (\eg IC15, SVTP and CUTE80). In overall performance on 7 benchmarks, our proposed GA-SPIN obviously surpasses all the existing methods under fair comparison.

\begin{figure}[!t]
	\centering
	\includegraphics[scale=0.5]{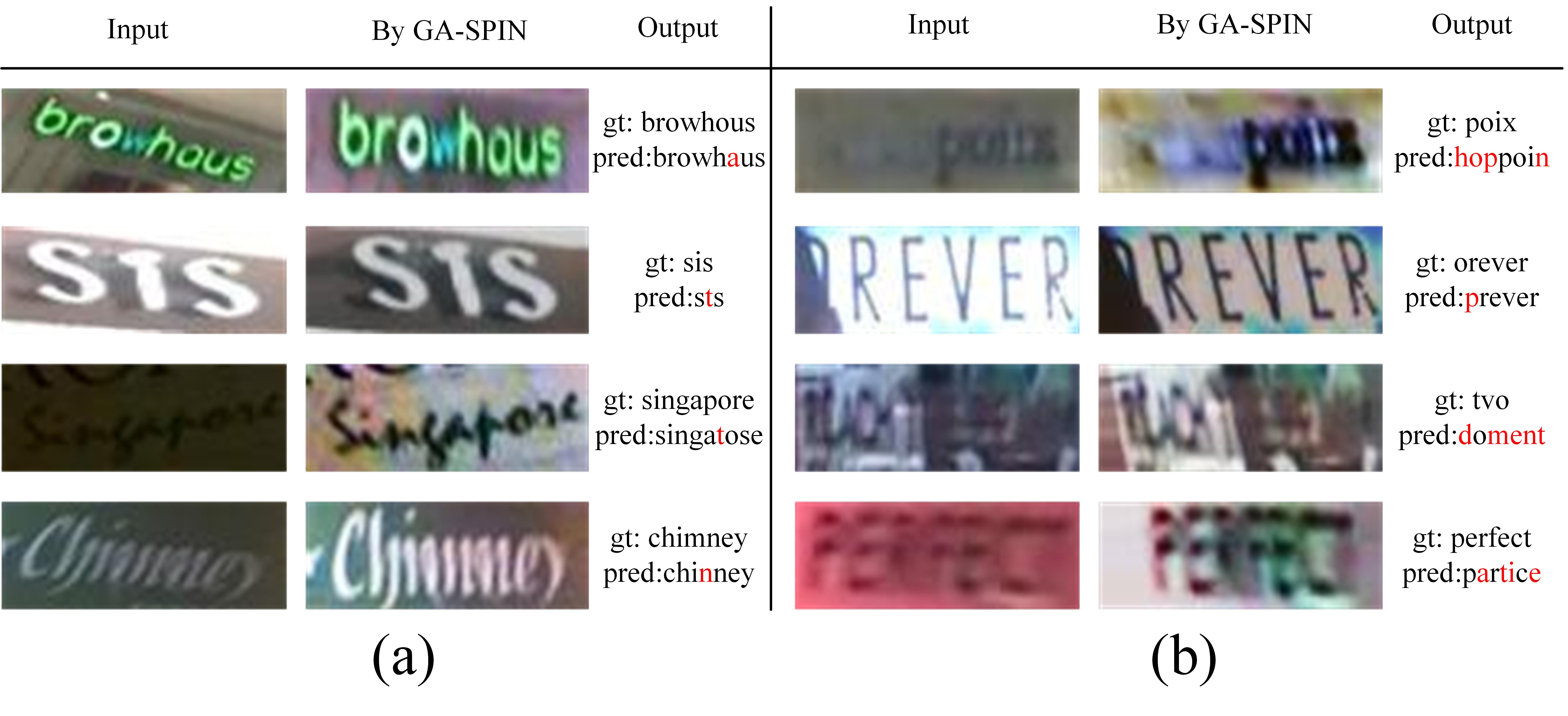}
    \vspace{-0.6cm}
	\caption{Some bad cases produced by our recognition system based on GA-SPIN. ``gt'' is the ground truth. ``pred'' is the prediction. Red are mistakenly recognized characters. (a) samples are from regular datasets, while sample in (b) are from irregular datasets. } %Samples are from SVT, IC15 and CUTE80.}
	\label{fig:supp7}
\end{figure}
\subsubsection{Failure Cases.}
Failure cases are shown in Fig. \ref{fig:supp7}. Similar characters like `o' and `a', `i' and `t' in the first two images are still challenging in recognition.
More difficulties include but not limit to shield (\eg ``orever'' in Fig. \ref{fig:supp7} (b)), stain (\eg ``poix''), blur (\eg ``perfect'') and severe complex background (\eg ``tvo'') in real world. These cannot be directly solved by such general yet flexible transformations like STN-based or our SPIN-based networks, left unsolved. % in the STR tasks. 
%%%%%%%%%%%%%%%%%%%%%%%%%%%%%%%5 conclusion%%%%%%%%%%%%%%%%%%%%%%%%%%%%%%%%%%%%%%%%%%%%%%%%
\section{Conclusion}
This paper proposes a novel idea of chromatic rectification in scene text recognition. The proposed SPIN allows the channel intensity manipulation of data within the network, giving neural networks the ability to actively transform the input color for clearer patterns.
Extensive experiments show its benefits the performance, especially in more complex scenes.
It is also verified that geometric and chromatic rectifications can be unified into GA-SPIN rather than pipeline of two modules, which further boosts the results and outperform the existing methods by a large margin.
%%%%%%%%%%%%%%%%%%%%%%%%%%%%%%%%%%%%%%%%%%%%%%%%%%%%%%%%%%%%%%%%%%%%%%%%%%%%%%%%%%%%%%%%%%%%%%%

\bibliography{bibfile}

\begin{thebibliography}{54}
\providecommand{\natexlab}[1]{#1}
\providecommand{\url}[1]{\texttt{#1}}
\providecommand{\urlprefix}{URL }
\expandafter\ifx\csname urlstyle\endcsname\relax
  \providecommand{\doi}[1]{doi:\discretionary{}{}{}#1}\else
  \providecommand{\doi}{doi:\discretionary{}{}{}\begingroup
  \urlstyle{rm}\Url}\fi

\bibitem[{Baek et~al.(2019)Baek, Kim, Lee, Park, Han, Yun, Oh, and Lee}]{wrong}
Baek, J.; Kim, G.; Lee, J.; Park, S.; Han, D.; Yun, S.; Oh, S.~J.; and Lee, H.
  2019.
\newblock What Is Wrong With Scene Text Recognition Model Comparisons? Dataset
  and Model Analysis.
\newblock In \emph{ICCV}, 4714--4722.

\bibitem[{Bhunia et~al.(2019)Bhunia, Das, Bhunia, Kishore, and Roy}]{AFDM}
Bhunia, A.~K.; Das, A.; Bhunia, A.~K.; Kishore, P. S.~R.; and Roy, P.~P. 2019.
\newblock Handwriting Recognition in Low-Resource Scripts Using Adversarial
  Learning.
\newblock In \emph{CVPR}, 4767--4776.

\bibitem[{Bookstein(1989)}]{TPS}
Bookstein, F.~L. 1989.
\newblock Principal warps: Thin-plate splines and the decomposition of
  deformations.
\newblock \emph{TPAMI} 11(6): 567--585.

\bibitem[{Chen et~al.(2020)Chen, Jin, Zhu, Luo, and Wang}]{Survey}
Chen, X.; Jin, L.; Zhu, Y.; Luo, C.; and Wang, T. 2020.
\newblock Text Recognition in the Wild: A Survey.
\newblock \emph{CoPR, abs/2005.03492} .

\bibitem[{Cheng et~al.(2017)Cheng, Bai, Xu, Zheng, Pu, and Zhou}]{FAN}
Cheng, Z.; Bai, F.; Xu, Y.; Zheng, G.; Pu, S.; and Zhou, S. 2017.
\newblock Focusing attention: Towards accurate text recognition in natural
  images.
\newblock In \emph{ICCV}, 5076--5084.

\bibitem[{Cheng et~al.(2018)Cheng, Xu, Bai, Niu, Pu, and Zhou}]{AON}
Cheng, Z.; Xu, Y.; Bai, F.; Niu, Y.; Pu, S.; and Zhou, S. 2018.
\newblock Aon: Towards arbitrarily-oriented text recognition.
\newblock In \emph{CVPR}, 5571--5579.

\bibitem[{Chung et~al.(2014)Chung, Gulcehre, Cho, and Bengio}]{GRU}
Chung, J.; Gulcehre, C.; Cho, K.; and Bengio, Y. 2014.
\newblock Empirical evaluation of gated recurrent neural networks on sequence
  modeling.
\newblock \emph{CoPR, abs/1412.3555} .

\bibitem[{Gupta, Vedaldi, and Zisserman(2016)}]{ST}
Gupta, A.; Vedaldi, A.; and Zisserman, A. 2016.
\newblock Synthetic data for text localisation in natural images.
\newblock In \emph{CVPR}, 2315--2324.

\bibitem[{He et~al.(2015)He, Zhang, Ren, and Jian}]{He}
He, K.; Zhang, X.; Ren, S.; and Jian, S. 2015.
\newblock Delving Deep into Rectifiers: Surpassing Human-Level Performance on
  ImageNet Classification.
\newblock In \emph{ICCV}, 1026--1034.

\bibitem[{He et~al.(2016)He, Zhang, Ren, and Sun}]{ResNet}
He, K.; Zhang, X.; Ren, S.; and Sun, J. 2016.
\newblock Deep Residual Learning for Image Recognition.
\newblock In \emph{CVPR}, 770--778.

\bibitem[{Hochreiter and Schmidhuber(1997)}]{LSTM}
Hochreiter, S.; and Schmidhuber, J. 1997.
\newblock Long short-term memory.
\newblock \emph{Neural Computation} 9(8): 1735--1780.

\bibitem[{Hu et~al.(2020)Hu, Cai, Hou, Yi, and Lin}]{GTC}
Hu, W.; Cai, X.; Hou, J.; Yi, S.; and Lin, Z. 2020.
\newblock GTC: Guided Training of CTC Towards Efficient and Accurate Scene Text
  Recognition.
\newblock In \emph{AAAI}, 11005--11012.

\bibitem[{Jaderberg et~al.(2014)Jaderberg, Simonyan, Vedaldi, and
  Zisserman}]{MJ}
Jaderberg, M.; Simonyan, K.; Vedaldi, A.; and Zisserman, A. 2014.
\newblock Synthetic data and artificial neural networks for natural scene text
  recognition.
\newblock \emph{CoPR, abs/1406.2227} .

\bibitem[{Jaderberg et~al.(2016)Jaderberg, Simonyan, Vedaldi, and
  Zisserman}]{JaderbergReading}
Jaderberg, M.; Simonyan, K.; Vedaldi, A.; and Zisserman, A. 2016.
\newblock Reading Text in the Wild with Convolutional Neural Networks.
\newblock \emph{IJCV} 116(1): 1--20.

\bibitem[{Jaderberg et~al.(2015)Jaderberg, Simonyan, Zisserman et~al.}]{STN}
Jaderberg, M.; Simonyan, K.; Zisserman, A.; et~al. 2015.
\newblock Spatial Transformer Networks.
\newblock In \emph{NeurIPS}, 2017--2025.

\bibitem[{Karatzas et~al.(2015)Karatzas, Gomez-Bigorda, Nicolaou, Ghosh,
  Bagdanov, Iwamura, Matas, Neumann, Chandrasekhar, Lu et~al.}]{IC15}
Karatzas, D.; Gomez-Bigorda, L.; Nicolaou, A.; Ghosh, S.; Bagdanov, A.;
  Iwamura, M.; Matas, J.; Neumann, L.; Chandrasekhar, V.~R.; Lu, S.; et~al.
  2015.
\newblock ICDAR 2015 competition on robust reading.
\newblock In \emph{ICDAR}, 1156--1160. IEEE.

\bibitem[{Karatzas et~al.(2013)Karatzas, Shafait, Uchida, Iwamura, i~Bigorda,
  Mestre, Mas, Mota, Almazan, and De~Las~Heras}]{IC13}
Karatzas, D.; Shafait, F.; Uchida, S.; Iwamura, M.; i~Bigorda, L.~G.; Mestre,
  S.~R.; Mas, J.; Mota, D.~F.; Almazan, J.~A.; and De~Las~Heras, L.~P. 2013.
\newblock ICDAR 2013 robust reading competition.
\newblock In \emph{ICDAR}, 1484--1493. IEEE.

\bibitem[{Landau, Smith, and Jones(1988)}]{landau1988importance}
Landau, B.; Smith, L.~B.; and Jones, S.~S. 1988.
\newblock The importance of shape in early lexical learning.
\newblock \emph{Cognitive development} 3(3): 299--321.

\bibitem[{Lee and Osindero(2016)}]{R2AM}
Lee, C.-Y.; and Osindero, S. 2016.
\newblock Recursive recurrent nets with attention modeling for ocr in the wild.
\newblock In \emph{CVPR}, 2231--2239.

\bibitem[{Li et~al.(2019)Li, Wang, Shen, and Zhang}]{2DATT}
Li, H.; Wang, P.; Shen, C.; and Zhang, G. 2019.
\newblock Show, Attend and Read: A Simple and Strong Baseline for Irregular
  Text Recognition.
\newblock In \emph{AAAI}, volume~33, 8610--8617.

\bibitem[{Liao et~al.(2019)Liao, Zhang, Wan, Xie, Liang, Lyu, Yao, and
  Bai}]{Perspective}
Liao, M.; Zhang, J.; Wan, Z.; Xie, F.; Liang, J.; Lyu, P.; Yao, C.; and Bai, X.
  2019.
\newblock Scene Text Recognition from Two-Dimensional Perspective.
\newblock In \emph{AAAI}, 8714--8721.

\bibitem[{Liu et~al.(2016)Liu, Chen, Wong, Su, and Han}]{STAR}
Liu, W.; Chen, C.; Wong, K.; Su, Z.; and Han, J. 2016.
\newblock STAR-Net: a SpaTial attention residue network for scene text
  recognition.
\newblock In \emph{BMVC}, 4168--4176.

\bibitem[{Liu et~al.(2018{\natexlab{a}})Liu, Wang, Jin, and Wassell}]{SSFL}
Liu, Y.; Wang, Z.; Jin, H.; and Wassell, I. 2018{\natexlab{a}}.
\newblock Synthetically supervised feature learning for scene text recognition.
\newblock In \emph{ECCV}, 435--451.

\bibitem[{Liu et~al.(2018{\natexlab{b}})Liu, Wang, Jin, and Wassell}]{Synth}
Liu, Y.; Wang, Z.; Jin, H.; and Wassell, I.~J. 2018{\natexlab{b}}.
\newblock Synthetically Supervised Feature Learning for Scene Text Recognition.
\newblock In \emph{ECCV}, 449--465.

\bibitem[{Lucas et~al.(2003)Lucas, Panaretos, Sosa, Tang, Wong, and
  Young}]{IC03}
Lucas, S.~M.; Panaretos, A.; Sosa, L.; Tang, A.; Wong, S.; and Young, R. 2003.
\newblock ICDAR 2003 robust reading competitions.
\newblock In \emph{ICDAR}, 682--687. Citeseer.

\bibitem[{Luo, Jin, and Sun(2019)}]{MORAN}
Luo, C.; Jin, L.; and Sun, Z. 2019.
\newblock MORAN: A Multi-Object Rectified Attention Network for Scene Text
  Recognition.
\newblock \emph{Pattern Recognition} 109--118.

\bibitem[{Luo et~al.(2020)Luo, Zhu, Jin, and Wang}]{Aug}
Luo, C.; Zhu, Y.; Jin, L.; and Wang, Y. 2020.
\newblock Learn to Augment: Joint Data Augmentation and Network Optimization
  for Text Recognition.
\newblock In \emph{CVPR}, 13743--13752.

\bibitem[{Lyu et~al.(2019)Lyu, Yang, Leng, Wu, Li, and Shen}]{lyu20192d}
Lyu, P.; Yang, Z.; Leng, X.; Wu, X.; Li, R.; and Shen, X. 2019.
\newblock 2D Attentional Irregular Scene Text Recognizer.
\newblock \emph{CoRR} abs/1906.05708.

\bibitem[{Mishra, Alahari, and Jawahar(2012)}]{IIIT5K}
Mishra, A.; Alahari, K.; and Jawahar, C. 2012.
\newblock Scene Text Recognition using Higher Order Language Priors.
\newblock In \emph{BMVC}, 1--11.

\bibitem[{Neumann and Matas(2012)}]{realtime}
Neumann, L.; and Matas, J. 2012.
\newblock Real-time scene text localization and recognition.
\newblock In \emph{CVPR}, 3538--3545.

\bibitem[{Peng, Zheng, and Zhang(2019)}]{SPT}
Peng, D.; Zheng, Z.; and Zhang, X. 2019.
\newblock Structure-Preserving Transformation: Generating Diverse and
  Transferable Adversarial Examples.
\newblock \emph{CoRR} abs/1809.02786.

\bibitem[{Quy~Phan et~al.(2013)Quy~Phan, Shivakumara, Tian, and Lim~Tan}]{SVTP}
Quy~Phan, T.; Shivakumara, P.; Tian, S.; and Lim~Tan, C. 2013.
\newblock Recognizing text with perspective distortion in natural scenes.
\newblock In \emph{ICCV}, 569--576.

\bibitem[{Risnumawan et~al.(2014)Risnumawan, Shivakumara, Chan, and
  Tan}]{CUTE80}
Risnumawan, A.; Shivakumara, P.; Chan, C.~S.; and Tan, C.~L. 2014.
\newblock A robust arbitrary text detection system for natural scene images.
\newblock \emph{Expert Systems with Applications} 41(18): 8027--8048.

\bibitem[{Shi, Bai, and Yao(2016)}]{CRNN}
Shi, B.; Bai, X.; and Yao, C. 2016.
\newblock An End-to-end Trainable Neural Network for Image-based Sequence
  Recognition and Its Application to Scene Text Recognition.
\newblock \emph{TPAMI} 39(11): 2298--2304.

\bibitem[{Shi et~al.(2016)Shi, Wang, Lyu, Yao, and Bai}]{RARE}
Shi, B.; Wang, X.; Lyu, P.; Yao, C.; and Bai, X. 2016.
\newblock Robust Scene Text Recognition with Automatic Rectification.
\newblock In \emph{CVPR}, 4168--4176.

\bibitem[{Shi et~al.(2019)Shi, Yang, Wang, Lyu, Yao, and Bai}]{ASTER}
Shi, B.; Yang, M.; Wang, X.; Lyu, P.; Yao, C.; and Bai, X. 2019.
\newblock ASTER: An Attentional Scene Text Recognizer with Flexible
  Rectification.
\newblock \emph{TPAMI} 41(9): 2035--2048.

\bibitem[{Simonyan and Zisserman(2015)}]{VGG}
Simonyan, K.; and Zisserman, A. 2015.
\newblock Very Deep Convolutional Networks for Large-Scale Image Recognition.
\newblock In \emph{ICLR}, 770--778.

\bibitem[{Su and Lu(2014)}]{su2014accurate}
Su, B.; and Lu, S. 2014.
\newblock Accurate scene text recognition based on recurrent neural network.
\newblock In \emph{ACCV}, 35--48. Springer.

\bibitem[{Wan et~al.(2020)Wan, He, Chen, Bai, and Yao}]{TextScanner}
Wan, Z.; He, M.; Chen, H.; Bai, X.; and Yao, C. 2020.
\newblock TextScanner: Reading Characters in Order for Robust Scene Text
  Recognition.
\newblock In \emph{AAAI}, 12120--12127.

\bibitem[{Wan et~al.(2019)Wan, Xie, Liu, Bai, and Yao}]{2DCTC}
Wan, Z.; Xie, F.; Liu, Y.; Bai, X.; and Yao, C. 2019.
\newblock 2D-CTC for Scene Text Recognition.
\newblock \emph{CoRR} abs/1907.09705.

\bibitem[{Wang and Hu(2017)}]{GRCNN}
Wang, J.; and Hu, X. 2017.
\newblock Gated Recurrent Convolution Neural Network for OCR.
\newblock In \emph{NeurIPS}, 335--344.

\bibitem[{Wang, Babenko, and Belongie(2011)}]{SVT}
Wang, K.; Babenko, B.; and Belongie, S. 2011.
\newblock End-to-end scene text recognition.
\newblock In \emph{ICCV}, 1457--1464.

\bibitem[{Wang and Belongie(2010)}]{wang-word}
Wang, K.; and Belongie, S. 2010.
\newblock Word spotting in the wild.
\newblock In \emph{ECCV}, 591--604. Springer.

\bibitem[{Wang et~al.(2020)Wang, Zhu, Jin, Luo, Chen, Wu, Wang, and Cai}]{DAN}
Wang, T.; Zhu, Y.; Jin, L.; Luo, C.; Chen, X.; Wu, Y.; Wang, Q.; and Cai, M.
  2020.
\newblock Decoupled Attention Network for Text Recognition.
\newblock In \emph{AAAI}, 12216--12224.

\bibitem[{Xie et~al.(2019)Xie, Huang, Zhu, Jin, Liu, and Xie}]{ACE}
Xie, Z.; Huang, Y.; Zhu, Y.; Jin, L.; Liu, Y.; and Xie, L. 2019.
\newblock Aggregation Cross-Entropy for Sequence Recognition.
\newblock In \emph{CVPR}, 6538--6547.

\bibitem[{Yang et~al.(2019)Yang, Guan, Liao, He, Bian, Bai, Yao, and
  Bai}]{ScRN}
Yang, M.; Guan, Y.; Liao, M.; He, X.; Bian, K.; Bai, S.; Yao, C.; and Bai, X.
  2019.
\newblock Symmetry-Constrained Rectification Network for Scene Text
  Recognition.
\newblock In \emph{ICCV}, 9147--9156.

\bibitem[{Yang et~al.(2017)Yang, He, Zhou, Kifer, and Giles}]{YangX}
Yang, X.; He, D.; Zhou, Z.; Kifer, D.; and Giles, C.~L. 2017.
\newblock Learning to Read Irregular Text with Attention Mechanisms.
\newblock In \emph{IJCAI}, 3280--3286.

\bibitem[{Yao et~al.(2016)Yao, Bai, Shi, and Liu}]{strokelets}
Yao, C.; Bai, X.; Shi, B.; and Liu, W. 2016.
\newblock Strokelets: {A} Learned Multi-Scale Mid-Level Representation for
  Scene Text Recognition.
\newblock In \emph{{IEEE} Trans. Image Process}, 2789--2802.

\bibitem[{Yu et~al.(2020)Yu, Li, Zhang, Han, and Ding}]{SRN}
Yu, D.; Li, X.; Zhang, C.; Han, J.; and Ding, E. 2020.
\newblock Towards Accurate Scene Text Recognition with Semantic Reasoning
  Networks.
\newblock In \emph{CVPR}, 12110--12119.

\bibitem[{Yue et~al.(2020)Yue, Kuang, Lin, Sun, and Zhang}]{RobustScanner}
Yue, X.; Kuang, Z.; Lin, C.; Sun, H.; and Zhang, W. 2020.
\newblock RobustScanner: Dynamically Enhancing Positional Clues for Robust Text
  Recognition.
\newblock In \emph{ECCV}.

\bibitem[{Zeiler(2012)}]{adadelta}
Zeiler, M.~D. 2012.
\newblock ADADELTA: an adaptive learning rate method.
\newblock \emph{CoPR, abs/1212.5701} .

\bibitem[{Zhan and Lu(2019)}]{ESIR}
Zhan, F.; and Lu, S. 2019.
\newblock {ESIR:} End-To-End Scene Text Recognition via Iterative Image
  Rectification.
\newblock In \emph{CVPR}, 2059--2068.

\bibitem[{Zhang et~al.(2020)Zhang, Yao, Yang, Xu, and Bai}]{AutoSTR}
Zhang, H.; Yao, Q.; Yang, M.; Xu, Y.; and Bai, X. 2020.
\newblock Efficient Backbone Search for Scene Text Recognition.
\newblock \emph{CoRR} abs/2003.06567.

\bibitem[{Zhang et~al.(2019)Zhang, Nie, Liu, Xu, Zhang, and Shen}]{SSDAN}
Zhang, Y.; Nie, S.; Liu, W.; Xu, X.; Zhang, D.; and Shen, H.~T. 2019.
\newblock Sequence-To-Sequence Domain Adaptation Network for Robust Text Image
  Recognition.
\newblock In \emph{CVPR}, 2740--2749.

\end{thebibliography}

\end{document}